\documentclass[]{zgca}

\usepackage{wrapfig}
\usepackage{tabularx}
\usepackage{textcomp}
\usepackage{stfloats}
\usepackage{url}
\usepackage{verbatim}
\usepackage{graphicx}
\usepackage{titlesec}
\usepackage{tocloft}
\usepackage{adjustbox}
\usepackage{multirow}
\usepackage{pifont}
\usepackage{tikz}
\usepackage{comment}
\usepackage{amsmath,amssymb} 
\usepackage{colortbl}  
\usepackage{color}
\usepackage{booktabs} 
\usepackage{hyperref}
\usepackage{graphicx}    
\usepackage{subcaption} 
\usepackage{booktabs}
\usepackage{amsmath} 
\usepackage{multirow} 
\usepackage{booktabs} 
\usepackage{subcaption} 
\usepackage{graphicx}   
\usepackage{wrapfig}    
\usepackage{caption}    
\usepackage{xparse}     

\RequirePackage{xspace}
\makeatletter
\DeclareRobustCommand\onedot{\futurelet\@let@token\@onedot}
\def\@onedot{\ifx\@let@token.\else.\null\fi\xspace}

\makeatother

\usepackage{makecell}

\definecolor{adptorange}{RGB}{248, 205, 172}
\definecolor{cmpblue}{RGB}{189, 215, 238}
\definecolor{cmpblue}{RGB}{189, 215, 238}

\definecolor{our_red}{RGB}{232,157,160}
\definecolor{our_blue}{RGB}{136,206,230}
\definecolor{our_orange}{RGB}{246,200,168}
\definecolor{our_green}{RGB}{178,211,164}

\definecolor{attn_code0}{RGB}{247,215,200}
\definecolor{attn_code1}{RGB}{238,169,139}
\definecolor{mlp_code0}{RGB}{204,201,221}
\definecolor{mlp_code1}{RGB}{102,95,153}

\definecolor{token_blue}{RGB}{84, 120, 140}

\definecolor{myMagenta}{rgb}{0.9,0,0.4}

\usepackage{pifont}       
\usepackage{bbding}       
\usepackage{fontawesome}
\usepackage{xspace}

\usepackage{float}

\newlength\savewidth

\newcolumntype{x}[1]{>{\centering\arraybackslash}p{#1pt}}
\newcolumntype{y}[1]{>{\raggedright\arraybackslash}p{#1pt}}
\newcolumntype{z}[1]{>{\raggedleft\arraybackslash}p{#1pt}}

\renewcommand{\paragraph}[1]{\vspace{1mm}\noindent\textbf{#1}}
\usepackage{colortbl}
\usepackage{xcolor}
\usepackage{wrapfig}

\renewcommand{\paragraph}[1]{\vspace{1.25mm}\noindent\textbf{#1}}

\usepackage{algorithm}
\usepackage{listings}

\definecolor{codeblue}{rgb}{0.25, 0.5, 0.5}
\definecolor{codekw}{rgb}{0.35, 0.35, 0.75}
\lstdefinestyle{Pytorch}{
    language = Python,
    backgroundcolor = \color{white},
    basicstyle = \fontsize{9pt}{8pt}\selectfont\ttfamily\bfseries,
    columns = fullflexible,
    aboveskip=1pt,
    belowskip=1pt,
    breaklines = true,
    captionpos = b,
    commentstyle = \color{codeblue},
    keywordstyle = \color{codekw},
}

\definecolor{green}{HTML}{009000}
\definecolor{red}{HTML}{ea4335}

\def\methodName{PhysBrain\xspace}
\def\datasetName{E2E-3M\xspace}

\title{\methodName: Human Egocentric Data as a Bridge from Vision Language Models to Physical Intelligence}

\author[1, 3, *]{Xiaopeng Lin}
\author[2, 6, *]{Shijie Lian}
\author[2, 5, *]{Bin Yu}
\author[4]{Ruoqi Yang}
\author[2]{Zhaolong Shen}
\author[2]{Changti Wu}
\author[2,5]{Yuzhuo Miao}
\author[3]{Yurun Jin}
\author[3]{Yukun Shi}
\author[2, 3]{Jiyan He}
\author[2, 3]{Cong Huang}
\author[\dagger 1]{Bojun Cheng}
\author[\dagger 2, 3, 4]{Kai Chen}

\affiliation[1]{The Hong Kong University of Science and Technology (Guangzhou)}
\affiliation[2]{Zhongguancun Academy}
\affiliation[3]{Zhongguancun Institute of Artificial Intelligence}
\affiliation[4]{DeepCybo}
\affiliation[5]{Harbin Institute of Technology}
\affiliation[6]{Huazhong University of Science and Technology}
\contribution[*]{Equal contribution}
\contribution[\dagger]{Corresponding author}

\abstract{

Robotic generalization relies on physical intelligence: the ability to reason about state changes, contact-rich interactions, and long-horizon planning under egocentric perception and action. Vision Language Models (VLMs) are essential to Vision–Language–Action (VLA) systems, but the reliance on third-person training data creates a viewpoint gap for humanoid robots. Collecting massive robot-centric data is an ideal but impractical solution due to cost and diversity constraints. Conversely, human egocentric videos offer a highly scalable data source with rich interaction context, yet the embodiment mismatch prevents the direct application. To bridge this gap, we propose an \textbf{Egocentric2Embodiment Translation Pipeline} that transforms raw human egocentric videos into multi-level, schema-driven embodiment supervision with enforced evidence grounding and temporal consistency, enabling the construction of the Egocentric2Embodiment dataset (\textbf{\datasetName}) at scale. An egocentric-aware embodied brain, termed \textbf{\methodName}, is obtained by training on the \datasetName dataset. \methodName exhibits substantially improved egocentric understanding, particularly for planning. It provides an egocentric-aware initialization that enables more sample-efficient VLA fine-tuning and higher success rates, demonstrating effective transfer from human egocentric supervision to downstream robot control.

}

\date{\today} 
\code{\url{https://zgc-embodyai.github.io/PhysBrain/}}
\correspondence{bocheng@hkust-gz.edu.cn, kaichen@zgci.ac.cn}

\begin{document}
\thispagestyle{firstheader}
\maketitle
\pagestyle{plain}

\let\cite\citep  

\makeatother

\section{Introduction}
\label{sec:introduction}




\begin{figure*}[!htb]
  \centering
   \includegraphics[width=1\linewidth]{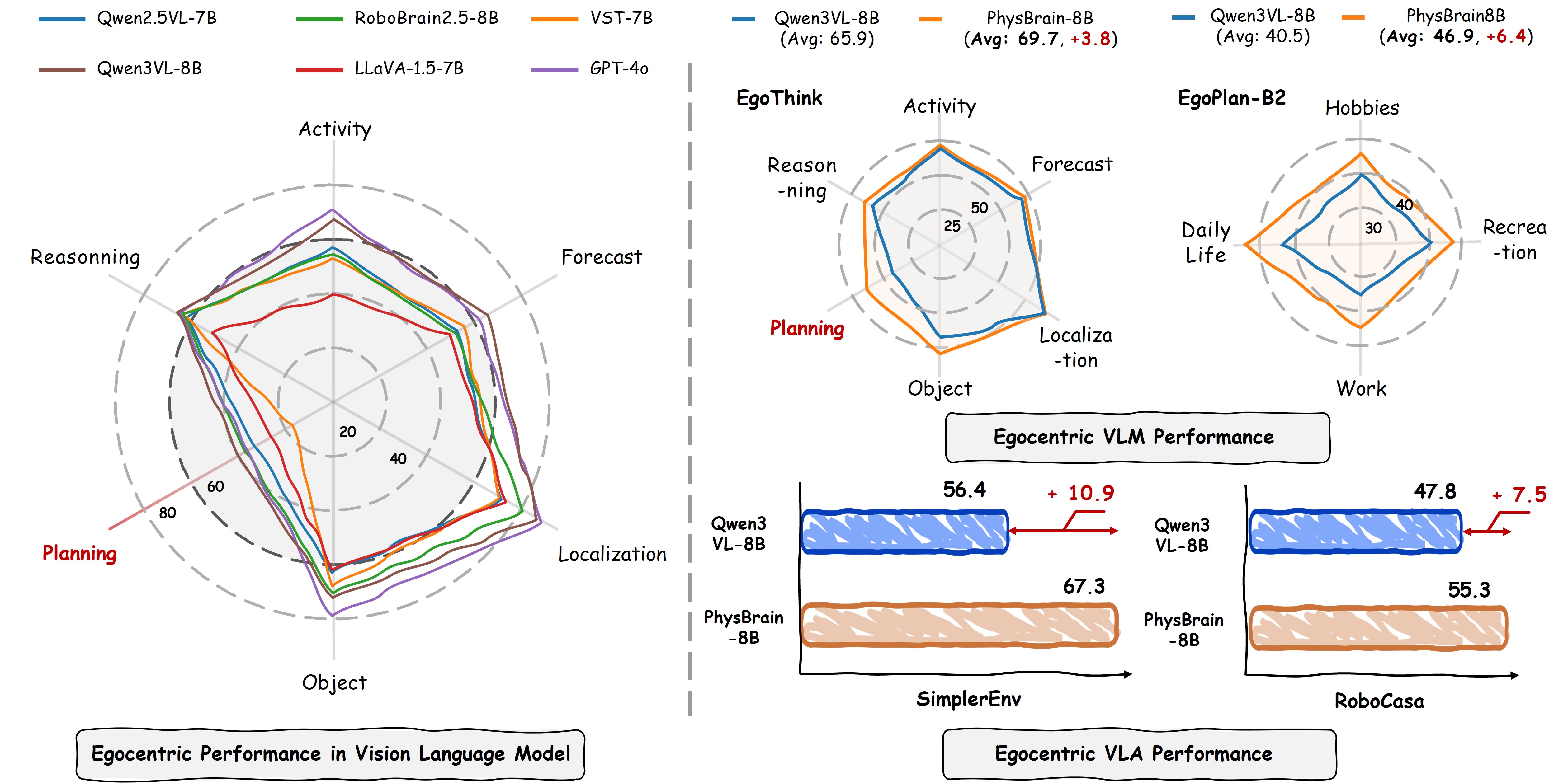}
  \caption{\textbf{Left}: EgoThink radar plot comparing egocentric VLM performance across six dimensions.
\textbf{Right Top}: The egocentric VLM performance on EgoThink and EgoPlan-B2 benchmarks with Qwen3VL-8B and PhysBrain, a Qwen3VL model fine-tuned with our annotated egocentric data (Sec.~\ref{subsec:Egocentric2Embodiment}). \textbf{Right Bottom}: The egocentric VLA performance on SimplerEnv and RoboCasa benchmarks as the VLM backbone in a VLA fine-tuning pipeline.}
   \label{fig:insight}
\end{figure*}


Vision-Language-Action (VLA) systems rely on a reliable embodied brain that integrates scenario understanding and action generation. Recent multimodal systems~\citep{GPT4o_2024_arXiv, Qwen3VL_2025_arXiv} show rapid gains in visual perception, spatial and video reasoning, and long context understanding. These advances provide rich open vocabulary recognition and semantic inference capabilities that can be transferred to action prediction, thereby enabling modern VLAs~\cite{RT2_2023_CoRL,OpenVLA_2024_CoRL,GR00TN1_2025_arXiv,pi0_2024_arXiv,pi05_2025_CoRL,VideoVLA_2025} to achieve strong performance across diverse manipulation tasks. These developments highlight that strong VLA performance is driven by an embodied brain that grounds executable planning and interaction decisions in the agent’s own perceptual stream.


For future humanoid robots, this perceptual stream is expected to be predominantly first-person, since perception, planning, and action feasibility are fundamentally grounded in the agent’s own body and workspace~\cite{Ego4D_2022_CVPR}. This places stringent demands on multimodal models operating under egocentric settings. However, empirical results on egocentric benchmarks~\cite{EgoVLP_2022_NIPS, EgoVLPv2_2023_ICCV, EgoPlanBench_2023_arXiv, QaEgo4Dv2_2025_arXiv} indicate that current multimodal models still struggle with long-horizon understanding, planning, and reliability under egocentric videos as shown in Fig.\ref{fig:insight}. These deficits stem from challenges intrinsic to egocentric perception, including rapid viewpoint changes, frequent hand–object occlusions, the absence of the actor’s full body, and the need for cross-frame inference of contact and object state~\cite{EgoReview_2025_arXiv}. Consequently, current performance bottlenecks are more likely due to insufficient egocentric embodied cognition, state tracking, and planning supervision, rather than limitations in model scale or single-frame recognition.


These limitations raise a fundamental scalability question: whether advancing VLA in egocentric settings necessarily depends on extensive robot data, including robot egocentric supervision. Acquiring large-scale and diverse robot manipulation data is widely acknowledged to be costly and difficult to scale, due to substantial hardware, labor, and safety constraints~\cite{khazatsky2024droid}. Even imitation learning relies on expensive human demonstrations, while existing large-scale robot data pipelines require long collection cycles or sustained multi-institution collaboration~\cite{brohan2022rt, RT2_2023_CoRL,o2024open}. As a result, learning and aligning embodied brains primarily through such robot data fundamentally constrains the scalability and coverage of egocentric VLA systems.


In contrast to costly and hard-to-scale robot data, human first-person videos provide a naturally scalable source of egocentric supervision, covering diverse everyday behaviors and environments. This data modality offers observations closely aligned with real interaction distributions for learning embodied brains. Large-scale datasets, such as Ego4D~\cite{Ego4D_2022_CVPR}, BuildAI~\cite{buildaiegocentric10k2025}, and EgoDex~\cite{EgoDex_2025_arXiv} demonstrate that egocentric videos can capture long-horizon activities, human–object interactions, and fine-grained manipulation dynamics at scale. An open question is how to leverage the latent planning structure and hand–object interaction regularities in human videos as supervision to strengthen egocentric embodied brains without robot data, thereby improving the sample efficiency and generalization of VLA systems.


Motivated by this observation, we develop a scalable annotation and instruction pipeline: Egocentric2Embodiment (E2E) Translation Pipeline that transforms human egocentric videos into structured, multi-level first-person VQA supervision for embodied brain learning. Each VQA instance encodes complementary information across multiple levels, including planning decompositions, key states, interaction constraints, and temporal relations, providing supervision beyond static visual recognition.



We instantiate the E2E translation pipeline at scale to construct E2E-3M from Ego4D, BuildAI, and EgoDex. The resulting supervision is used to train \methodName over the Qwen3-VL backbone. As shown in Fig.\ref{fig:insight}, \methodName substantially outperforms the base model on the Egocentric VLM benchmarks, with the clearest improvements in egocentric perception and planning. When plugged into the egocentric VLA system as the embodied brain, \methodName continues to get excellent performance on the VLA benchmarks. Few-shot adaptation on SimplerEnv and RoboCasa is effective and sample-efficient. The resulting performance exceeds VLA systems trained with large-scale robot data, while using no robot-data pretraining. Comprehensive evaluation results across VLM and VLA benchmarks validate E2E-3M dataset and the E2E Translation Pipeline, supporting human egocentric data as an effective source of physically grounded supervision for embodied brains. In summary, our contributions are as follows:
\begin{itemize}
    \item We introduce a scalable annotation and instruction pipeline, called \textbf{Egocentric2Embodiment Translation Pipeline}, which converts large-scale human egocentric videos into multi-level embodied supervision.


    \item We provide a well-structured and validated egocentric VQA dataset \textbf{\datasetName{}} that can effectively improve models' first-person vision performance and generalization capability on VLA tasks.

    \item Extensive experiments have demonstrated that human egocentric videos provide effective supervision for learning embodied brains in egocentric settings, leading to improved generalization in VLA tasks.

\end{itemize}
\section{Related Work}
\label{sec:related_work}


\subsection{First-Person Vision Language Model}


Vison Language Models (VLMs) that excel on third-person content often degrade when the input shifts to egocentric imagery and video. 
Multiple lines of evidence point to a persistent viewpoint domain gap and to missing egocentric cues such as hand manipulation, egomotion, and partial observability \cite{EgoReview_2025_arXiv}.
EgoVLP~\cite{EgoVLP_2022_NIPS} is among the first to document that third-person pretraining transfers poorly and that explicitly egocentric objectives are needed for first-person retrieval, recognition, and temporal grounding.
EgoVLPv2~\cite{EgoVLPv2_2023_ICCV} further reports that fusing first-person video and language during pretraining is important for egocentric tasks.
Beyond these early works,
EgoPlan-Bench~\cite{EgoPlanBench_2023_arXiv} shows that mainstream multimodal models struggle with egocentric planning even when the scenes are household and the instructions are simple, and it analyzes typical failure modes such as viewpoint confusion and missing contact reasoning.
Studies on QaEgo4D~\cite{QaEgo4D_2022_CVPRW} and QaEgo4Dv2 \cite{QaEgo4Dv2_2025_arXiv} find that both proprietary and open source VLMs lag on long-horizon egocentric reasoning.
EgoM2P~\cite{EgoM2P_2025_arXiv} also emphasizes the structural gap between third-person and first-person streams and argues for egocentric priors during pretraining.

\subsection{Vision Language Action}


Vision-Language-Action (VLA) models~\citep{rt1_2022,RT2_2023_CoRL,Octo_2024,pi0_2024_arXiv,lian2026langforcebayesiandecompositionvision,yu2026twinbrainvla} represent a recent paradigm shift in robotic manipulation by unifying language understanding, visual perception, and motor control within a single end-to-end framework. Building upon large-scale vision-language models, VLAs directly map high-dimensional visual observations and natural language instructions to low-level robot actions, enabling intuitive human-robot interaction and task execution. Early works such as RT-1~\citep{rt1_2022} and RT-2~\citep{RT2_2023_CoRL} demonstrate that scaling robot data and leveraging pretrained vision-language representations significantly improve manipulation performance across diverse tasks. Building upon these foundations, OpenVLA~\citep{OpenVLA_2024_CoRL}, $\pi_0$~\citep{pi0_2024_arXiv,pi-FAST_2025,pi05_2025_CoRL}, and GR00T-N1~\citep{GR00TN1_2025_arXiv} further advance VLA capabilities through large-scale cross-embodiment and multi-task pretraining, demonstrating superior generalization and action prediction performance. Several works~\citep{ChatVLA2_2025_nips,InstructVLA_2025,DualVLA_2025,HyT_2025} attempt to address the catastrophic forgetting of language capabilities during VLA training, while others~\citep{ECoT_2024,EmmaX_2024,OneTwoVLA_2025,ThinkAct_2025,MolmoAct_2025,EmbodiedR1_2025} explore incorporating chain-of-thought reasoning into the VLA inference process. To pursue better generalization, several works~\citep{VideoVLA_2025,MMACT_2025,Video2Act_2025} attempt to incorporate video generation models or world models into VLA action prediction, while others~\citep{SimpleVLA-RL_2025,RLinf_2025,piRL_2025,TGRPO_2025} explore applying reinforcement learning to train VLA models. 


\subsection{Learning VLAs from Human Demonstration}


Robot data acquisition is hard to scale due to the stringent robot–operator configuration and reliance on expert tele-operation. Egocentric VLA trained on the egocentric human demonstrations offers a more scalable path, with strong potential to advance perception–action learning and real-world executability. EgoVLA~\citep{EgoVLA_2025} utilizes scaled egocentric videos plus a unified human–robot action space with light robot finetuning, enabling efficient skill transfer. Being-H0~\citep{BeingH0_2025} leverages physical-instruction tuning with discrete hand-motion codes and a physics-aligned cross-view space supports fine-grained VLA training from human videos. H-RDT~\citep{H-RDT_2025} sets large bimanual pretraining with 3D hand pose and a two-stage diffusion policy delivers substantial improvements. GR-3~\citep{GR3_2025} utilizes multi-source training yields strong generalization, rapid few-shot adaptation, and robust long-horizon bimanual and mobile control.  RynnVLA-001~\citep{jiang2025rynnvla} pretrains on large-scale human video demonstrations with video generation objectives and compresses actions into a continuous latent space via ActionVAE to align video prediction with downstream robot fine-tuning. VITRA~\citep{VITRA_2025} treats the human hand as a proxy end-effector, converts the egocentric hand videos into robot-aligned formats, and combines VLMs with diffusion-based action experts for policy learning.

However, these approaches focus on the explicit alignment of human data to robot action spaces, which is inherently constrained by embodiment gaps. In contrast, our work targets a more upstream objective by transforming egocentric human data into embodiment supervision signals for embodied brains, providing a scalable foundation that complements robot-data-based pipelines.

\section{Egocentric Embodied Supervision}

In this section, we introduce the egocentric data annotation pipeline and the \textbf{\datasetName} dataset.

\subsection{Egocentric2Embodiment Translation Pipeline}
\label{subsec:Egocentric2Embodiment}

\begin{figure*}[ht!]
  \centering
   \includegraphics[width=1\linewidth]{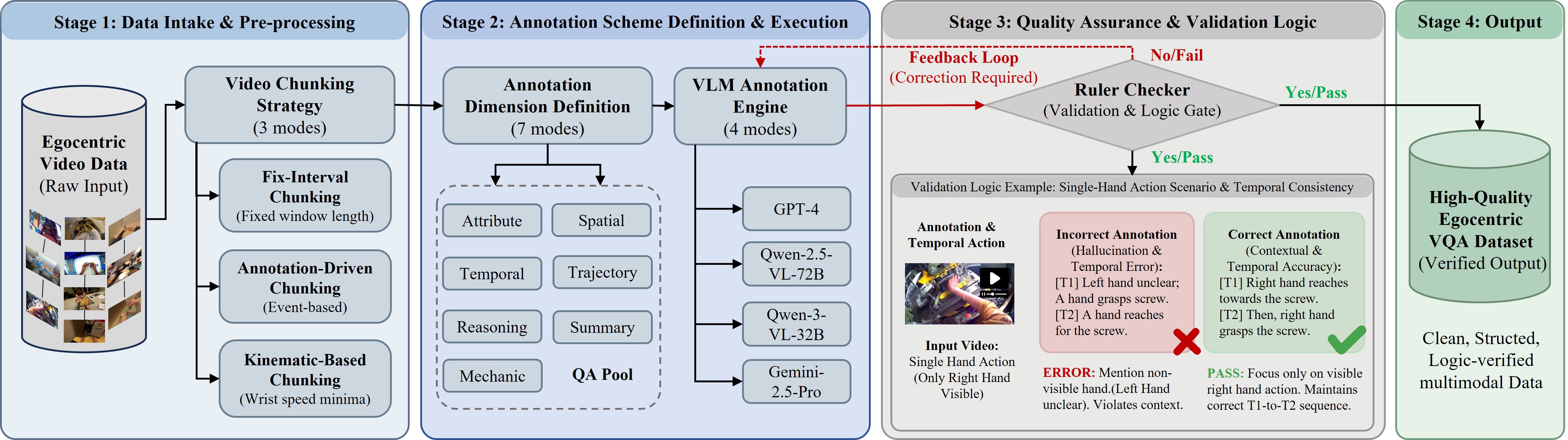}
   \caption{Illustration of the Egocentric2Embodiment Translation Pipeline.}
   \label{fig:dataset}
\end{figure*}

Human egocentric videos encode rich embodied experience, including action progression, hand–object interaction, and task-level structure. However, this experience is not directly usable for training embodied brains. Raw videos lack explicit structure, free-form language annotations are unstable, and unconstrained generation often introduces temporal ambiguity or hallucinated interactions.

Our key idea is to translate human egocentric data into structured and verifiable supervision that captures the hierarchical structure of embodied behavior, spanning action semantics, temporal organization, interaction dynamics, and task-level reasoning. To this end, we design a schema-driven, rule-validated egocentric VQA data engine as shown in Fig.\ref{fig:dataset} that systematically converts raw egocentric human videos into multi-level supervision aligned with embodied planning and interaction reasoning.

\subsubsection{Data Intake and Pre-processing}
To define the basic supervision units, the engine chunks each episode into short temporal clips at stage 1. Given the large variation in egocentric action amplitude and frequency across scenarios, we adopt scenario-aware temporal segmentation, including fixed-interval, event-driven\cite{Egocentric10k_2025_HuggingFace}, and kinematic-aware strategies\cite{VITRA_2025}. All clips are associated with explicit temporal spans and exposed through a unified interface for downstream annotation. Episode-level metadata is used as contextual conditioning to limit the semantic space of subsequent question answering. 

The resulting representations are temporally localized and preserve short-range state transitions relevant to embodied manipulation and interaction. It forms the atomic unit for the downstream annotation stage.

\subsubsection{Annotation Scheme}

In stage 2, we define a finite, schema-driven annotation space to produce supervision that reflects embodied cognition rather than generic video description. Each clip is labeled with one of seven complementary VQA modes, including temporal, spatial, attribute, mechanics, reasoning, summary, and trajectory. Each mode is paired with a template set that standardizes wording and controls the information granularity. The engine samples a mode and a template, then generates a customized and detailed QA pair for each clip.

VQA generation is performed by a set of VLM annotation engines. The schema constrains both the question form and the required semantic content, which keeps supervision targets consistent across different generators. Answers must be natural-language and grounded in the visual evidence. The engine enforces egocentric conventions such as left/right hand references and manipulation-specific phrasing. This stage yields multi-level annotations that capture complementary aspects of planning and interaction reasoning.

\subsubsection{Quality Assurance and Validation Logic}

Open-ended generation easily produces errors that are harmful for training supervision. Common failures include references to non-visible hands and incorrect temporal ordering, et al. We therefore introduce the third stage that a rule checker is designed as a validation gate. Samples that fail validation are rejected and sent back for regeneration with a structured error that indicates the violated constraint.


The checker applies three types of constraints designed to filter visual hallucinations. Evidence grounding restricts the generation scope, requiring that mentioned entities are grounded in the episode-level object metadata. Egocentric consistency enforces the correct hand references and prohibits mentions of unseen limbs or contradictory assignments. Mode-specific temporal logic requires explicit temporal connectors and verifies timeline alignment. The generation–validation loop repeats until all constraints are satisfied. To further guarantee the visual fidelity of the resulting dataset, we conducted a rigorous human audit on a random subset, confirming that our logic-verified filtering effectively retains high-quality, hallucination-free supervision.

\subsubsection{Structured Egocentric Corpus}

Samples that satisfy all validation constraints are retained and compiled into the egocentric VQA supervision dataset. Each entry records the sampled frames, the selected VQA mode, the generated QA pair, and the validation outcome. This design ensures traceability and reproducibility. The supervision produced by the proposed data engine offers structured and logic-verified supervision that encodes action organization and hand–object interaction, completing the translation of egocentric video data into reliable training signals for egocentric planning and interaction reasoning.

\subsection{Egocentric2Embodiment Dataset (E2E-3M)}

\subsubsection{Data Sources and Domain Coverage}

The proposed Egocentric2Embodiment Translation Pipeline is applied to generate the \datasetName, a large-scale human egocentric video corpora collected across three complementary domains: household, factory, and laboratory environments. These corpora captures substantial variation in environmental context, object composition, and interaction patterns, as shown in Fig.\ref{fig:dataset_overview} and Fig.\ref{fig:dataset_sample}.

Specifically, Ego4D represents open-world household activities and provides extensive geographic and contextual diversity. BuildAI captures real industrial workflows, emphasizing procedural regularity and dense hand visibility in factory environments. EgoDex focuses on laboratory settings and offers high-resolution egocentric manipulation sequences with fine-grained interaction cues. These sources differ systematically in spatial layout, object distribution, and task structure. The aggregation yields the \datasetName dataset with complementary coverage across the space of egocentric embodied experience.

\begin{figure}[t!]
  \centering
   \includegraphics[width=1\linewidth]{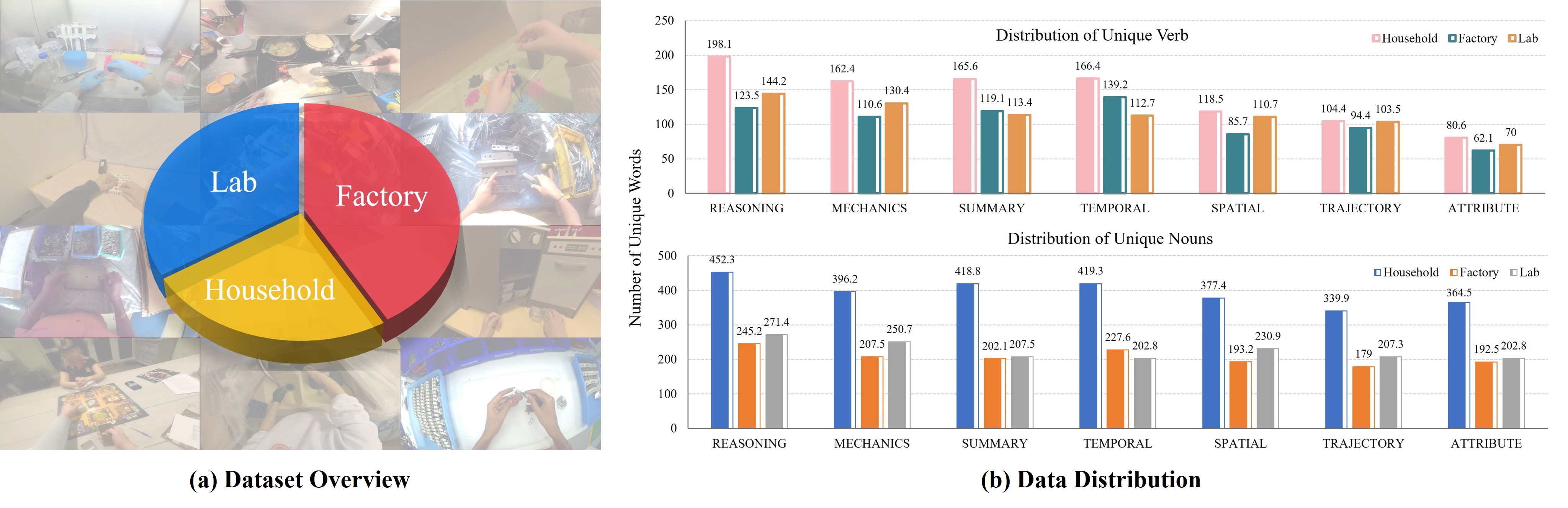}
   \caption{Data Distribution Statistics of \datasetName{} dataset.}
   \label{fig:dataset_analysis}
\end{figure}

\subsubsection{Diversity Analysis}

To evaluate whether the dataset provides sufficiently rich supervision for generalized embodied manipulation, we analyze diversity along two interpretable axes: object coverage and action (verb) coverage, as quantified in Eq.~\ref{eq:obj_div} and Eq.~\ref{eq:verb_div}.

\noindent\textbf{Object Coverage and Environmental Spectrum.} 
Object coverage reflects the breadth of perceptual contexts. We calculate the normalized noun lemma count per domain  as:
\begin{equation}
\label{eq:obj_div}
\text{ObjectDiv}(s) = \frac{|\mathcal{V}^{\text{noun}}_s|}{T^{\text{noun}}_s} \times 1000,
\end{equation}
where $|\mathcal{V}^{\text{noun}}_s|$ is the number of unique noun lemmas and $T^{\text{noun}}_s$ is the total noun token count in domain $s$.
The distribution shown in Fig.\ref{fig:dataset_analysis}(b) validates that our aggregated dataset spans the complete spectrum from structured to unstructured environments. 
Specifically, \textit{Household} data exhibits high object diversity, capturing the long-tail distribution of objects typical in unstructured, open-world scenarios. \textit{Factory} data shows lower ObjectDiv scores, which accurately reflect the standardized, repetitive nature of industrial workflows where agents interact with a fixed set of tools and parts. \textit{Lab} settings fall in between, offering a semi-structured middle ground. This distributional shift is not a limitation but a critical feature: it ensures that VLA models are trained on both the rigid procedural constraints of industry and the chaotic variability of daily life, fostering robust generalization.

\noindent\textbf{Action Coverage and Interaction Semantics.} 
Action coverage quantifies the richness of manipulation semantics. We calculate the verb diversity as:
\begin{equation}
\label{eq:verb_div}
\text{VerbDiv}(m) = \frac{|\mathcal{V}^{\text{verb}}_m|}{N_m} \times 1000,
\end{equation}
where normalization by the number of QA pairs $N_m$ accounts for interaction density. As shown in Fig.~\ref{fig:dataset_analysis}, action-centric modes (e.g., Reasoning, Mechanics, Temporal) consistently maintain high lexical diversity across all domains. Notably, even in the \textit{Factory} domain where object diversity is lower due to standardization, the \textit{Mechanics} and \textit{Reasoning} scores remain high. This decoupling indicates that while the industrial environment is visually structured, the underlying manipulation tasks involve complex, non-trivial procedural logic. This confirms that our dataset provides dense supervision for manipulation behaviors regardless of the environmental structure.

The \datasetName{} dataset bridges human egocentric video and embodied brain learning by providing structured supervision with broad scene coverage and rich action diversity. We expect that releasing this dataset will support future research on egocentric VLA and physical intelligence.
\section{Methodology}
\label{sec:method}

\begin{wrapfigure}{l}{0.5\textwidth}  
  \centering
  \includegraphics[width=0.95\linewidth]{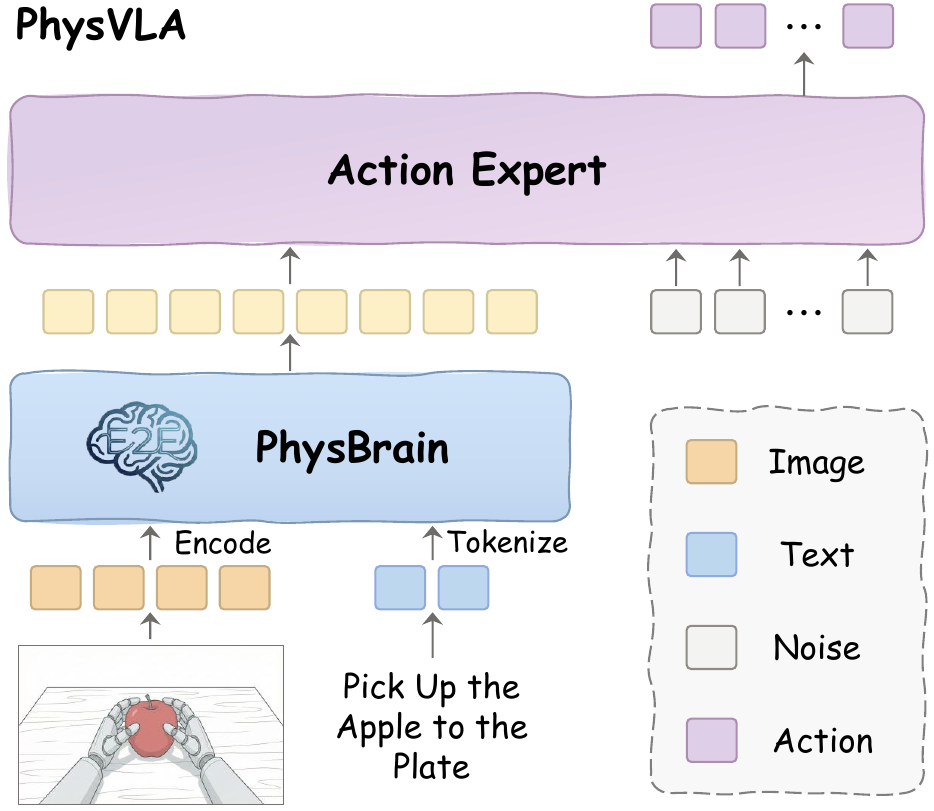}  
  \caption{\textbf{VLA architecture built on \methodName}. PhysVLA conditions a flow-matching diffusion action expert on the \emph{last-layer} hidden states of \methodName.}
  \label{fig:framework}
\end{wrapfigure}

Using the data annotation pipeline proposed in the previous section, we translate embodied experience from egocentric videos into structured supervision suitable for learning an embodied brain. To preserve general-purpose vision--language capability during SFT, we additionally mix an equal-sized subset sampled from FineVision, a large-scale curated vision--language corpus. We then perform supervised fine-tuning (SFT) on base VLMs (e.g., Qwen3-VL-4B and Qwen3-VL-8B) using this mixture, resulting in an egocentric-centered VLM backbone (\methodName) with improved first-person understanding, reasoning, and planning capabilities. Quantitative results are reported in Table \ref{tab:combined_benchmark}.

With \methodName in hand, we study how these egocentric gains transfer to downstream control under standard VLA instantiations. Our goal in this section is not to propose a new VLA architecture, but to evaluate transferability while minimizing confounding factors from additional heuristics or hand-crafted priors. We follow the widely adopted community paradigm, GR00T-style, and keep the action expert lightweight and consistent.

We denote an observation (the egocentric image sequence) as $o_t$, the language instruction as $x$, and the VLM parameters as $\phi$. The VLM produces token-level hidden states:
\begin{equation}
\mathbf{H}^\ell_t = \mathrm{VLM}_\phi(o_t, x)[\ell] \in \mathbb{R}^{N \times d}, \quad \ell=1,\dots,L,
\end{equation}
where $L$ is the number of layers in the VLM, $N$ is the token length, and $d$ is the hidden dimension. The action policy predicts a future action chunk $\mathbf{a}_{t:t+K} \in \mathbb{R}^{K \times d_a}$.



\paragraph{PhysVLA.}
We introduce PhysVLA, which follows the dual-system design in GR00T N1.5~\cite{GR00TN1_2025_arXiv}: the VLM  plays the role of System~2 to produce high-level multimodal representations, while a Flow-Matching (FM) action expert~\cite{RectifiedFlow_2025_arXiv} serves as System~1 to generate continuous actions. Concretely, the \emph{last-layer} VLM hidden states $\mathbf{Z}_t = \mathbf{H}^{L}_t$~ are utilized as the conditioning signal.

The FM expert is implemented as a diffusion transformer~(DiT)~\cite{DiT_2023_ICCV} that denoises an action trajectory by cross-attending to $\mathbf{Z}_t$ (VLM features are keys/values, action tokens are queries). Under the rectified-flow parameterization, we sample Gaussian noise $\boldsymbol\epsilon \sim \mathcal{N}(0,\mathbf{I})$ and a time scalar $\tau \in (0,1]$, then linearly interpolate between noise and the target action chunk to obtain the noised trajectory $\tilde{\mathbf{a}}$:
\begin{equation}
\tilde{\mathbf{a}} = (1-\tau)\,\boldsymbol\epsilon + \tau\,\mathbf{a}, \qquad \mathbf{v} = \mathbf{a} - \boldsymbol\epsilon.
\end{equation}
Here $\mathbf{v}$ is the target (time-independent) velocity that transports the noise trajectory to the data trajectory under this parameterization. The action expert predicts this velocity field conditioned on VLM features (and optional proprioceptive state $\mathbf{s}_t$):
\begin{equation}
\hat{\mathbf{v}} = f_\theta(\tilde{\mathbf{a}}, \tau;\, \mathbf{Z}_t, \mathbf{s}_t),
\end{equation}
and is trained with a simple regression objective
\begin{equation}
\mathcal{L}_{\mathrm{FM}} = \mathbb{E}\big[\lVert \hat{\mathbf{v}} - \mathbf{v} \rVert_2^2\big].
\end{equation}
At inference, we start from noise and apply a small number of FM denoising steps (we use $\text{steps}=8$) to obtain the action chunk $\mathbf{a}_{t:t+K}$ with $K{=}16$. This design provides a controlled setting to examine how informative the egocentric VLM representation $\mathbf{Z}_t$ is for action prediction.


\definecolor{navyblue}{HTML}{0071BC}

\begin{table*}[ht!]
    \centering
    \caption{\textbf{Comparison on EgoPlan and EgoThink benchmarks.} The detailed sub-tasks belong exclusively to the EgoThink benchmark, with the overall average reported in the final column.}
    \vspace{0.5em}
    
    \resizebox{\textwidth}{!}{
        \begin{tabular}{l | c | c | ccccccc | c}
            \toprule
            
            \multirow{2.5}{*}{\textbf{Method}} & \textbf{EgoPlan-B1} & 
            \textbf{EgoPlan-B2} & 
            \multicolumn{8}{c}{\textbf{EgoThink Benchmark}} \\
            
            \cmidrule(lr){2-2} \cmidrule(lr){3-11} 
            
             & \textbf{Acc.} & \textbf{Acc.} & Act. & Fore. & Loc. & Obj. & Asst. & Nav. & Reas. & \textbf{Avg.} \\
            \midrule
            
            \rowcolor{navyblue!10}\multicolumn{11}{c}{\textit{\textbf{General VLM}}} \\
            \addlinespace[0.2em]
            
            GPT-4o~\citep{GPT4o_2024_arXiv}    & 39.5  & 41.0 & 73.0 & 66.0 & 89.0 & 78.6 & 28.0 & 12.0 & 66.0 & 66.4 \\
            MiniGPT-4-7B~\citep{zhu2023minigpt} & 28.1 & 24.5 & 45.5 & 36.5 & 61.5 & 48.0 & 30.0 & 12.0 & 36.7 & 41.6 \\
            LLaVA-1.5-7B~\citep{liu2024improved} & 27.8 & 25.4 & 35.0 & 43.5 & 76.0 & 65.3 & 33.0 & 26.0 & 53.0 & 51.6 \\
            LLaMA-3.2-11B~\citep{dubey2024llama} & 24.3 & 25.1   & 34.0 & 49.5 & 57.5 & 62.7 & 42.0 & 22.0 & 47.7 & 48.4 \\
            Qwen-3-VL-4B~\citep{yang2025qwen3} & 42.2 & 34.6 & 63.5 & 65.0 & 82.5 & 72.6 & 46.0 & 35.0 & 71.0 & 66.7 \\
            Qwen-3-VL-8B~\citep{yang2025qwen3} & 44.3 & 40.5 & 68.0 & 66.5 & 86.0 & 72.3 & 41.0 & 39.0 & 61.7 & 65.9 \\
            
            \rowcolor{navyblue!10}\multicolumn{11}{c}{\textit{\textbf{Embodied Brain}}} \\
            \addlinespace[0.2em]
            
            VST-RL-7B~\citep{VST_2025}      & 40.8    & 28.7 & 55.0 & 56.5 & 69.5 & 67.3 & 15.0 & 22.0 & 62.3 & 56.2 \\
            RoboBrain2.0-7B~\citep{RoboBrain2_2025} & 38.6 & 23.3 & 35.0 & 47.0 & 77.5 & 60.7 & 44.0 & 38.0 & 52.3 & 52.8 \\
            RoboBrain2.5-8B~\citep{tan2026robobrain25depthsight} & \underline{45.9} & \underline{45.2} & 57.5 & 56.5 & 81.0 & 70.3 & 40.0 & 28.0 & 68.3 & 62.4 \\

            \cmidrule(lr){1-11} 
            
            \textbf{PhysBrain-4B} (ours) & 43.9 & 39.3 & 68.0 & 64.5 & 85.5 & 76.3 & 66.0 & 44.0 & 66.0 & \underline{69.4} \\
            \textbf{PhysBrain-8B} (ours) & \textbf{47.4} & \textbf{46.9} & 69.0 & 69.0 & 86.5 & 76.0 & 65.0 & 42.0 & 64.0 & \textbf{69.7} \\
            \bottomrule
        \end{tabular}
    }
    \vspace{0.3em}
    \label{tab:combined_benchmark}
\end{table*}
\section{Experiment}
\label{sec:experiment}


This section details the experimental setup, benchmarks, and results. We report results from VLM and VLA evaluations.

\subsection{VLM Egocentric Evaluation}
\label{subsec:exper:vlm_eval}

\subsubsection{Egocentric Understanding Evaluation}


For a comprehensive egocentric evaluation, we conduct experiments on three widely-used benchmarks: EgoPlan-Benchmark1\cite{EgoPlanBench_2023_arXiv}, EgoPlan-Benchmark2\cite{qiu2024egoplan}, and EgoThink~\citep{cheng2024egothink}. These benchmarks cover diverse real-world settings and complementary protocols: EgoPlan utilizes multiple-choice evaluation, whereas EgoThink scores free-form generations via an LLM judge.

\textbf{EgoPlan-Benchmark1:} EgoPlan-Benchmark1\cite{EgoPlanBench_2023_arXiv} evaluates the planning capabilities of Vision Language Models in real-world scenarios from an egocentric perspective, simulating human perception. This benchmark includes realistic tasks involving a diverse range of action plans and intricate visual observations. We validate the VLM models using 3,314 VQA pairs from the EgoPlan-Val set, provides a comprehensive assessment of a model’s ability to understand task progress, track current states, and predict the next feasible action. EgoPlan-Benchmark1 presents significant challenges for current VLMs, highlighting substantial room for improvement in human-level task planning.

\textbf{EgoPlan-Benchmark2:} EgoPlan-Benchmark2\cite{qiu2024egoplan} extends the scope of its predecessor by incorporating a broader variety of real-world scenarios across four key domains: Work, Daily Life, Hobbies, and Recreation, with 24 distinct scenarios. This expanded benchmark includes 1,321 high-quality multiple-choice questions sourced from 1,113 egocentric videos, each representing a real-world task. The dataset emphasizes task planning, requiring models to predict the next action based on task progress, current visual observations, and task goals. EgoPlan-Benchmark2 provides a rigorous evaluation of VLMs' planning capabilities and highlights the challenges these models face in real-world decision-making, with detailed analysis and instructions to enhance planning performance.

\textbf{EgoThink:} EgoThink\cite{cheng2024egothink} evaluates the first-person perspective capabilities of vision-language models (VLMs) across six core areas: Object, Activity, Localization, Reasoning, Forecasting, and Planning. It contains 700 images from 595 egocentric videos annotated with question-answer pairs. GPT-4 is used as an automatic evaluator to grade open-ended responses. The EgoThink benchmark provides a comprehensive assessment of VLMs' reasoning and planning abilities in real-world scenarios from a first-person viewpoint, revealing key challenges and offering insights for advancing embodied AI.



\definecolor{navyblue}{HTML}{0071BC}


\begin{table*}[ht!]
  \centering
  \caption{
    \textbf{Results of evaluating the VLA models with the WidowX robot in the SimplerEnv simulation environment}. We highlight the best results in \textbf{bold} and the second-best results with \underline{underline}.
    }
  \begin{adjustbox}{width=\linewidth}
  \begin{tabular}{l c c c c c}
    \toprule
    \textbf{Method}
     & \makecell[c]{\textbf{Put Spoon} \\ \textbf{on Towel}} 
     & \makecell[c]{\textbf{Put Carrot} \\ \textbf{on Plate}} 
     & \makecell[c]{\textbf{Stack Green Block} \\ \textbf{on Yellow Block}} 
     & \makecell[c]{\textbf{Put Eggplant} \\ \textbf{in Yellow Basket}} 
     & \textbf{Average} \\
    \midrule

    \rowcolor{navyblue!10}\multicolumn{6}{c}{\textit{\textbf{VLA Baselines}}} \\
    \addlinespace[0.2em] 

    RT-1-X~\citep{o2024open}         &  0.0  & 4.2   & 0.0   & 0.0   & 1.1 \\
    Octo-Base~\citep{Octo_2024}       & 15.8  & 12.5  & 0.0   & 41.7  & 17.5 \\
    Octo-Small~\citep{Octo_2024}      & 41.7  & 8.2   & 0.0   & 56.7  & 26.7 \\
    OpenVLA~\citep{OpenVLA_2024_CoRL}         & 4.2   & 0.0   & 0.0   & 12.5   & 4.2 \\
    OpenVLA-OFT~\citep{OpenVLA-OFT_2025}     & 12.5  & 4.2  & 4.2  & 72.5  & 23.4 \\
    RoboVLM~\citep{RoboVLM_2024}         & 50.0  & 37.5  & 0.0   & 83.3  & 42.7 \\ 
    TraceVLA~\citep{TraceVLA_2025}        & 12.5  & 16.6  & 16.6  & 65.0  & 27.7 \\
    SpatialVLA~\citep{SpatialVLA_2025}      & 20.8  & 20.8  & 25.0  & 70.8  & 34.4 \\
    CogACT~\citep{CogACT_2024}          & 71.7 &  50.8  & 15.0 & 67.5 & 51.3 \\
    VideoVLA~\citep{VideoVLA_2025}        & 75.0 & 20.8   & 45.8 & 70.8 & 53.1 \\
    $\pi_0$~\citep{pi0_2024_arXiv}         & 29.1 & 0.0 & 16.6 & 62.5 & 27.1 \\
    $\pi_{0.5}$~\citep{pi05_2025_CoRL} & 49.3 & 64.7 & 44.7 & 69.7 & 57.1 \\
    Isaac-GR00T-N1.6-Bridge~\citep{GR00T_N1.6}   & 64.5 & 65.5 & 5.5 & 93.0 & 57.1 \\
    
    \cmidrule(lr){1-6}  
    
    \rowcolor{navyblue!10}\multicolumn{6}{c}{\textit{\textbf{VLM Baselines}}} \\
    \addlinespace[0.2em]

    Qwen2.5-VL-7B~\citep{Qwen25VL_2025_arXiv}       & 68.7 &  35.4 & 25.0 & 75.0 & 51.0 \\
    Qwen3-VL-4B~\citep{yang2025qwen3}  & 87.5 &  50.0 & 29.2 & 54.2 & 55.2 \\
    Qwen3-VL-8B~\citep{yang2025qwen3}  & 68.7 & 38.5 & 30.2 & 87.9 & 56.3 \\
    VST-RL-7B~\citep{VST_2025}           & 57.7 & 41.7 & 16.7 & 50.0 & 41.3 \\
    RoboBrain2.0-7B~\citep{RoboBrain2_2025}     & 30.8 & 24.7 & 2.5 & 93.3 & 37.8 \\
    RoboBrain2.5-8B~\citep{tan2026robobrain25depthsight}     & 75.0 & 55.5 & 40.1 & 100.0 & \textbf{67.6} \\
    
    \cmidrule(lr){1-6}  

    \textbf{PhysBrain-4B} (ours) & 90.3 & 58.3 & 34.7 &  80.6  & 65.9 \\
    \textbf{PhysBrain-8B} (ours) & 77.8 & 62.5 & 34.8 &  94.8  & \underline{67.4} \\

    \bottomrule
  \end{tabular}
  \end{adjustbox}
  \vspace{0.5 em}
  \label{tab:simplerenv_gr00t_results}
\end{table*}

\textbf{Baselines.} We primarily compare our method against two categories of baselines: (i) \textbf{General VLM}, which include closed-source models such as GPT-4o and widely-used open-source models (MiniGPT-4-7B, LLaVA-1.5-7B, LLaMA-3.2-11B, Qwen3-VL-4B, and Qwen3-VL-8B); and (ii) \textbf{Embodied Brain}, which include VST-RL-7B~\citep{VST_2025}, RoboBrain2-7B~\citep{RoboBrain2_2025} and RoboBrain2.5-8B~\citep{tan2026robobrain25depthsight}.

\textbf{Evaluation.} The comparison methods are evaluated through the released weight for direct inference. Evaluation conditions are standardized across models, and all models use the same prompt template as shown in Table \ref{tab:prompts}. These controls ensure that performance differences reflect model capability rather than prompt variation or inconsistent scoring.


In the Egothink benchmark, the generation outputs are scored with a single GPT-4~\citep{hurst2024gpt} judging protocol across all EgoThink subtasks. Table~\ref{tab:combined_benchmark} summarizes performance on the six EgoThink dimensions (Activity, Forecast, Localization, Object, Planning, Reasoning). PhysBrain-8B achieves the highest average performance, while our PhysBrain-4B achieves sub-optimal performance and consistently outperforms strong open and competitive baselines. The most pronounced improvement is observed on Planning, where PhysBrain substantially exceeds all baselines, indicating a clear advantage in translating egocentric observations into executable plans.

In the EgoPlan benchmark, we report accuracy on both EgoPlan-Benchmark1 and EgoPlan-Benchmark2 (Table~\ref{tab:combined_benchmark}). Trained with our E2E-3M supervision, PhysBrain consistently improves over the Qwen3-VL backbone: PhysBrain-8B reaches 47.4/46.9 on B1/B2, outperforming Qwen3-VL-8B (44.3/40.5) by +3.1/+6.4 points; PhysBrain-4B achieves 43.9/39.3, improving over Qwen3-VL-4B (42.2/34.6) by +1.7/+4.7. These gains support that E2E-3M injects physically grounded interaction priors into the base model, leading to stronger egocentric planning.


\subsection{VLA Simulation Evaluation}
\label{subsec:exper:vla_sim_eval}

\begin{table*}[h]
    \centering
    \small
    \renewcommand{\arraystretch}{1.3} 
    \setlength{\tabcolsep}{5pt} 

    \caption{
      \textbf{Results of evaluating the VLA models with the GR1 robot in the RoboCasa Tabletop simulation environment}. The results for QwenGR00T, QwenOFT, and QwenFAST are derived from the official StarVLA experiments~\citep{starvla_2025}. We highlight the best results in \textbf{bold} and the second-best results with \underline{underline}.
    }
    \resizebox{\textwidth}{!}{
    \begin{tabular}{l c c c c | c c}
        \toprule
        \textbf{Task} & 
        \textbf{\scriptsize \makecell{Isaac-GR00T\\N1.6}} & 
        \textbf{\scriptsize \makecell{QwenGR00T\\ + Qwen3VL}} &  
        \textbf{\scriptsize \makecell{QwenOFT\\ + Qwen3VL}} & 
        \textbf{\scriptsize \makecell{QwenFAST\\ + Qwen3VL}} & 
        \textbf{\scriptsize \makecell{\methodName{}-4B}} & 
        \textbf{\scriptsize \makecell{\methodName{}-8B}} \\
        \midrule
        PnP Bottle To Cabinet Close & 51.5 & 46.0 & 30.0 & 38.0 & 74.0 & 70.0 \\
        PnP Can To Drawer Close & 13.0 & 80.0 & 76.0 & 44.0 & 68.0 & 74.0 \\
        PnP Cup To Drawer Close & 8.5 & 54.0 & 44.0 & 56.0 & 42.0 & 46.0 \\
        PnP Milk To Microwave Close & 14.0 & 48.0 & 44.0 & 44.0 & 54.0 & 60.0 \\
        PnP Potato To Microwave Close & 41.5 & 28.0 & 32.0 & 14.0 & 24.0 & 34.0 \\
        PnP Wine To Cabinet Close & 16.5 & 46.0 & 36.0 & 14.0 & 54.0 & 40.0 \\
        PnP Novel From Cuttingboard To Basket & 58.0 & 48.0 & 50.0 & 54.0 & 62.0 & 54.0 \\
        PnP Novel From Cuttingboard To Cardboardbox & 46.5 & 40.0 & 40.0 & 42.0 & 44.0 & 56.0 \\
        PnP Novel From Cuttingboard To Pan & 68.5 & 68.0 & 70.0 & 58.0 & 56.0 & 72.0 \\
        PnP Novel From Cuttingboard To Pot & 65.0 & 52.0 & 54.0 & 58.0 & 58.0 & 74.0 \\
        PnP Novel From Cuttingboard To Tieredbasket & 46.5 & 56.0 & 38.0 & 40.0 & 40.0 & 44.0 \\
        PnP Novel From Placemat To Basket & 58.5 & 42.0 & 32.0 & 36.0 & 42.0 & 58.0 \\
        PnP Novel From Placemat To Bowl & 57.5 & 44.0 & 58.0 & 38.0 & 56.0 & 56.0 \\
        PnP Novel From Placemat To Plate & 63.0 & 48.0 & 52.0 & 42.0 & 80.0 & 62.0 \\
        PnP Novel From Placemat To Tieredshelf & 28.5 & 18.0 & 24.0 & 18.0 & 14.0 & 28.0 \\
        PnP Novel From Plate To Bowl & 57.0 & 60.0 & 60.0 & 52.0 & 54.0 & 70.0 \\
        PnP Novel From Plate To Cardboardbox & 43.5 & 50.0 & 50.0 & 30.0 & 50.0 & 54.0 \\
        PnP Novel From Plate To Pan & 51.0 & 54.0 & 66.0 & 48.0 & 68.0 & 56.0 \\
        PnP Novel From Plate To Plate & 78.7 & 70.0 & 68.0 & 50.0 & 78.0 & 60.0 \\
        PnP Novel From Tray To Cardboardbox & 51.5 & 38.0 & 44.0 & 28.0 & 40.0 & 52.0 \\
        PnP Novel From Tray To Plate & 71.0 & 56.0 & 56.0 & 34.0 & 66.0 & 60.0 \\
        PnP Novel From Tray To Pot & 64.5 & 50.0 & 62.0 & 46.0 & 52.0 & 70.0 \\
        PnP Novel From Tray To Tieredbasket & 57.0 & 36.0 & 54.0 & 36.0 & 50.0 & 48.0 \\
        PnP Novel From Tray To Tieredshelf & 31.5 & 16.0 & 30.0 & 16.0 & 22.0 & 28.0 \\
        \midrule
        \rowcolor{gray!30} 
        \textbf{Average} & 47.6 & 47.8 & 48.8 & 39.0 & \underline{49.75} & \textbf{55.25} \\
        \bottomrule
    \end{tabular}
    }
    \label{tab:robocasa_main_tab}
\end{table*}

To validate the efficacy of our model when deployed as the VLA for robotic control, we adopt \methodName{} as the VLM backbone and fine-tune it within the VLA paradigm using downstream robotics data. We then evaluate on the SimplerEnv~\citep{SimpleEnv_2024_CoRL} and RoboCasa~\citep{RoboCasa_2024_RSS} simulation benchmarks.

\textbf{SimplerEnv:} The WidowX Robot is utilized in  the SIMPLER environment to evaluate the VLA policy on four manipulation tasks: "Put Spoon on Towel," "Put Carrot on Plate," "Stack Green Block on Yellow Block," and "Put Eggplant in Yellow Basket." For each task, we conduct 24 trials and using the official evaluation script from the SimplerEnv repository. This ensures a consistent and rigorous assessment of the robot's manipulation performance across these tasks.

\textbf{RoboCasa:} RoboCasa GR1 Tabletop Manipulation Benchmark includes 24 diverse tasks involving complex interactions with articulated objects and varied geometries. Examples of tasks include "PnPBottleToCabinetClose" and "PnPCanToDrawerClose", as well as scenarios with appliances like microwaves and toasters. For training, we use the Humanoid Robot Tabletop Manipulation subset from the PhysicalAI-Robotics-GR00T-X-Embodiment-Sim dataset \citep{GR00T_2025_arXiv}, ensuring a robust and diverse training environment for evaluating robotic manipulation performance.

\subsubsection{Experiment Settings}


\textbf{Training.} To adapt the VLM to the VLA architecture and the target robotic platform, we follow the training configuration of the starVLA~\citep{starvla_2025} framework. Detailed training hyperparameters and configurations are provided in Appendix~\ref{sec:appendix:vla_bench_details}.


\textbf{Baselines.} We primarily compare our method against two categories of baselines: (i) \textbf{VLA baselines}, which include several widely used VLA models (RT-1-X, Octo, OpenVLA, RoboVLM, TraceVLA, SpatialVLA, CogACT, VideoVLA, $\pi_0$, $\pi_{0.5}$, and Issac-GR00T-N1.6-Bridge); and (ii) \textbf{VLM baselines}, where we fine-tune several commonly used VLMs (Qwen2.5-VL, Qwen3-VL, RoboBrain2.0, VST-RL, and RoboBrain2.5) under the PhysVLA paradigm and evaluate by using the same training configuration as ours.

\textbf{Evaluation.} We evaluate our policy on SimplerEnv~\citep{SimpleEnv_2024_CoRL} (four WidowX tasks) and RoboCasa (24 complex manipulation tasks involving articulated objects). To ensure statistical robustness, we adhere to official protocols, reporting mean success rates (Avg@50) averaged over 5 and 50 independent trials for SimplerEnv and RoboCasa, respectively.






\subsubsection{Experiment Results}

Table ~\ref{tab:simplerenv_gr00t_results} summarizes the SimplerEnv evaluation results, comparing our \methodName{} with all baseline methods. The evaluation results of RoboCasa are presented in Table \ref{tab:robocasa_main_tab}.


As detailed in Table~\ref{tab:simplerenv_gr00t_results}, PhysBrain-8B achieves an average success rate of 67.4\%, significantly surpassing VLA baselines trained on substantially larger robot datasets (e.g., Isaac-GR00T at 57.1\%) and performing comparably to the state-of-the-art RoboBrain2.5 (67.6\%). Crucially, while RoboBrain2.5 relies on massive cross-embodiment robot data for representation alignment, PhysBrain attains this proficiency solely through pretraining on our proposed \datasetName dataset. This highlights the potential of human data by providing robust physical priors that transfer seamlessly to robotic manipulation. 



We evaluate PhysBrain on the RoboCasa benchmark against the Qwen3-VL-4B backbone integrated with three distinct action encodings (GR00T, OFT, and FAST), as shown in Table~\ref{tab:robocasa_main_tab}. Since PhysBrain adopts the Gr00T architecture, the comparison with the standard QwenGR00T baseline is particularly critical, as it directly isolates the performance gains attributable to our proposed method. As shown in Table~\ref{tab:robocasa_main_tab}, PhysBrain-4B achieves a 49.75\% success rate, not only surpassing its direct counterpart (47.8\%) but also outperforming other encoding variants. Furthermore, the PhysBrain-8B model validates significant scalability, attaining the best average success rate of 55.25\%.

\subsection{Ablation Study}

\begin{wraptable}{l}{0.5\textwidth}  
    \centering
    \caption{Ablation experiment on the impact of the proposed E2E-3M dataset across VLM Architectures and Model Scales. Best results are in \textbf{bold}.}
    \label{tab:ablation_scale}
    
    \begin{tabular}{lcc}
        \toprule
        \textbf{Method} & \textbf{VLM Bench} & \textbf{VLA Bench} \\
        \midrule
        Qwen2.5-VL-7B   & 58.7 & 34.4 \\
        \textbf{PhysBrain2.5-7B (ours)}        & \textbf{63.1} & \textbf{53.9} \\
        \midrule
        Qwen3-VL-4B    & 66.7 & 55.2 \\
        \textbf{PhysBrain-4B (ours)}        & \textbf{69.4} & \textbf{65.9} \\
        \midrule
        Qwen3-VL-8B    & 65.9 & 56.3 \\
        \textbf{PhysBrain-8B (ours)}        & \textbf{69.7} & \textbf{67.4} \\
        \bottomrule
    \end{tabular}
\end{wraptable}

\textbf{Dataset Impact Across VLM Architectures and Model Scales.} 
To verify the universality of the proposed \datasetName dataset, we conducted extensive ablation studies across different VLM architectures (Qwen2.5-VL and Qwen3-VL) and varying parameter scales (4B, 7B, and 8B). As detailed in Table~\ref{tab:ablation_scale}, fine-tuning on our dataset consistently yields significant performance improvements over the base models across all configurations. Notably, while the VLM benchmark scores show steady increments, the improvements on the VLA benchmark are particularly substantial (e.g., a 19.5\% gain for the 7B model). This disparity indicates that while standard VLMs possess strong general vision-language capabilities, they lack the specific physical priors required for robotic manipulation. These results confirm that the physical intelligence distilled from \datasetName is agnostic to the underlying architecture, consistently enhancing the embodied planning capabilities of diverse foundation models.

\begin{wraptable}{l}{0.5\textwidth}  
    \centering
    \caption{Ablation experiment on the impact of the egocentric embodied supervision scale. Best results are in \textbf{bold}.}
    \label{tab:ablation_data_scale}
    
    \begin{tabular}{lcc}
        \toprule
        \textbf{Method} & \textbf{VLM Bench} & \textbf{VLA Bench} \\
        \midrule
        PhysBrain-8B-wo-Ego4D   & 67.8 & 64.1 \\
        \textbf{PhysBrain-8B (ours)}        & \textbf{69.7} & \textbf{67.4} \\
        \bottomrule
    \end{tabular}
\end{wraptable}

\textbf{Impact of Egocentric Supervision Data Scale.} 
We further investigate the scaling behavior of our approach by analyzing the impact of dataset size on model performance. Table~\ref{tab:ablation_data_scale} presents an ablation study where we remove the large-scale Ego4D subset from the training data. The results demonstrate a clear positive correlation between the volume of egocentric supervision and the resulting model proficiency. The full PhysBrain-8B model outperforms the version trained without Ego4D on both VLM and VLA benchmarks. This decline in performance upon data reduction highlights that the scale and diversity of the dataset are critical factors. It suggests that scaling up high-quality egocentric human data is a viable and effective pathway for continuously improving the physical understanding and manipulation skills of embodied agents.
\section{Conclusion}
\label{sec:conclusion}

In this work, we leverage human egocentric videos to connect vision-language models with physical intelligence for robotic generalization. We introduce the Egocentric2Embodiment pipeline, which converts raw egocentric videos into multi-level VQA supervision, creating the \datasetName{} dataset with nearly three million verified instances across household, factory, and laboratory settings. By fine-tuning a vision-language model on the \datasetName{} dataset, we develop \methodName{} that significantly enhances egocentric capabilities. It also achieves high success rates on VLA system when used as the VLM backbone in standard VLA fine-tuning. Our results demonstrate that scalable human egocentric supervision bridges vision-language understanding and physical intelligence, enabling greater data diversity and improved policy learning from human demonstrations.





\bibliographystyle{assets/plainnat}
\bibliography{ego}
\newpage
\beginappendix


\section{Dataset Overview}
\label{sec:appendix:Dataset_Overview}

\subsection{Dataset Composition}
\label{subsec:appendix:Dataset_Composition}


\begin{figure*}[h!]
  \centering
   \includegraphics[width=1\linewidth]{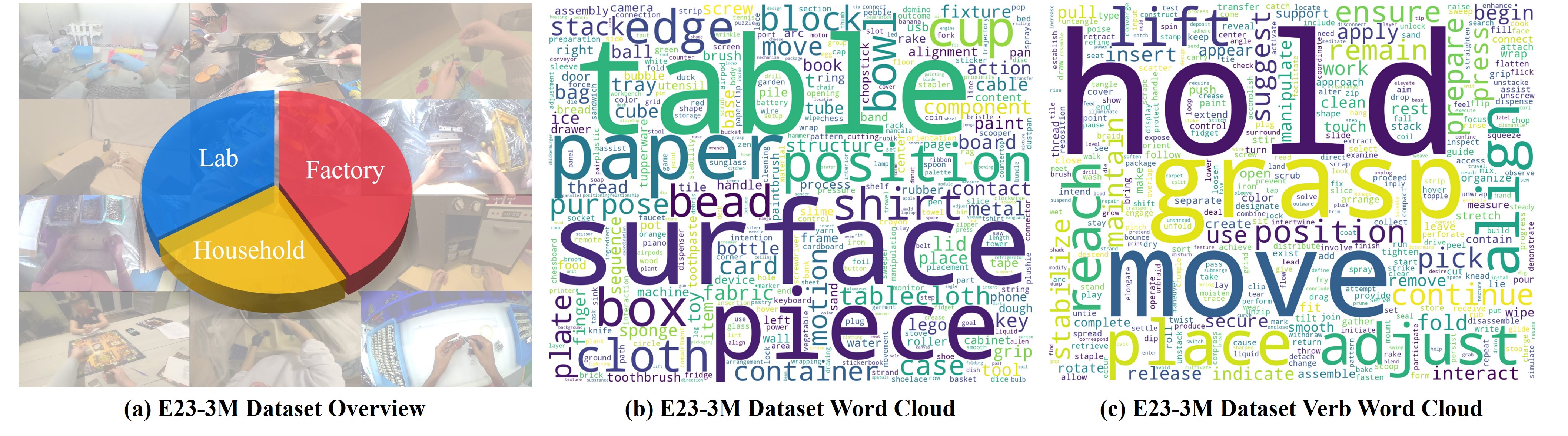}
   \caption{Overview and word cloud of \datasetName{} dataset.}
   \label{fig:dataset_overview}
\end{figure*}

\begin{figure*}[h!]
  \centering
   \includegraphics[width=1\linewidth]{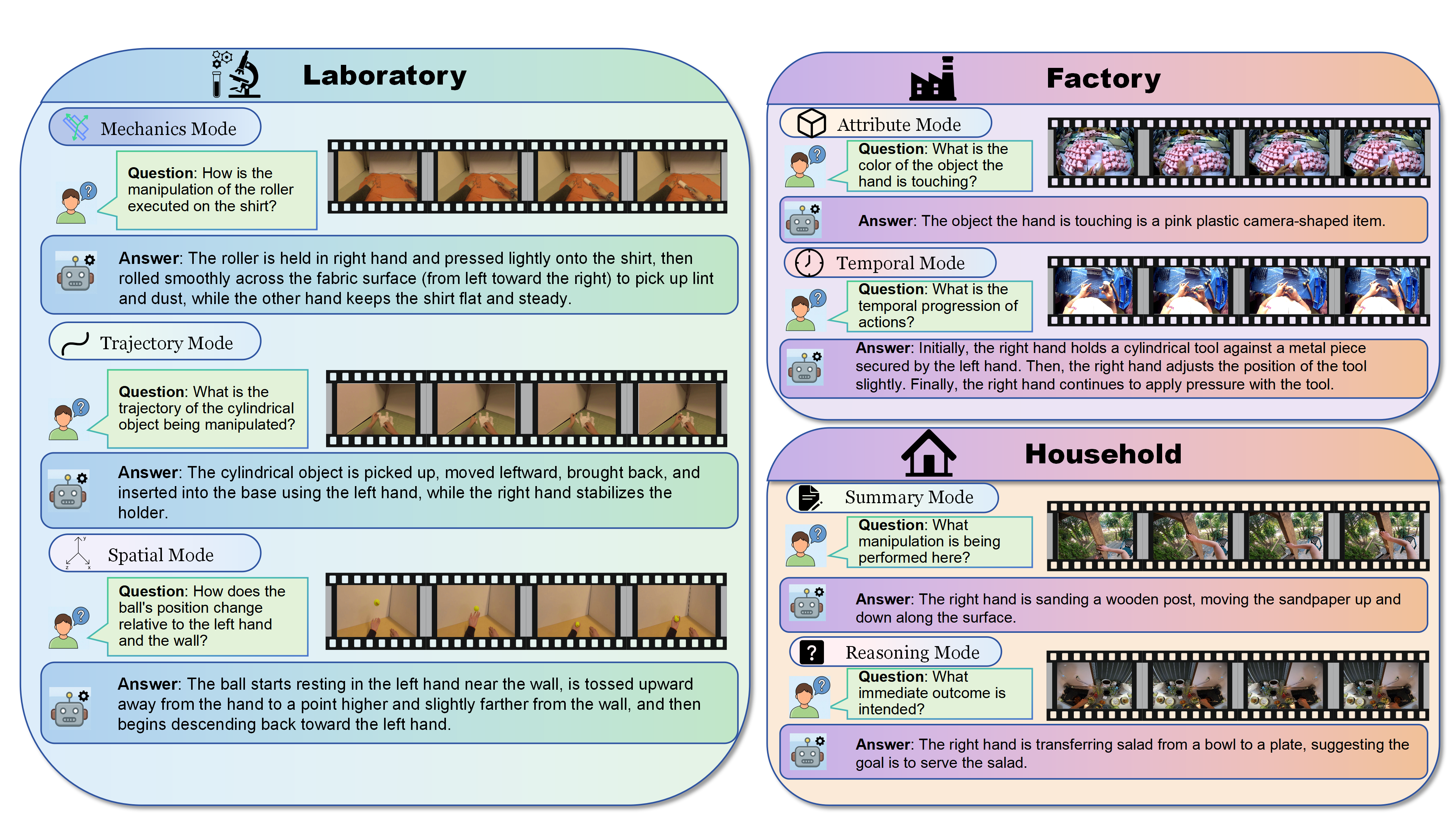}
   \caption{ Selected Examples of \datasetName{} dataset. All scenarios consist of seven modes.}
   \label{fig:dataset_sample}
\end{figure*}

The \datasetName dataset consists of approximately three million verified instances across three domains: Household, Factory, and Laboratory, as shown in Fig.\ref{fig:dataset_overview}(a). Each domain includes diverse scenarios where human egocentric data is collected and annotated to provide rich supervision for training vision-language-action models. The dataset is designed to cover a wide range of object manipulations and interactions in varied environments, offering valuable insights for improving VLA model performance.

\subsection{Verb and Noun Distribution}
\label{subsec:appendix:Dataset_Distribution}

The word clouds shown in the Fig.\ref{fig:dataset_overview} represent the most frequent nouns and verbs across the entire dataset. Fig.\ref{fig:dataset_overview}(b) highlights key nouns such as "table," "surface," and "box," reflecting the primary objects involved in the manipulations across different tasks. Fig.\ref{fig:dataset_overview}(c) shows the most common verbs, such as "hold," "grasp," and "move," which are central to the actions performed in the dataset. These word clouds provide a visual summary of the object and action diversity within the dataset, crucial for training models to recognize and act on a wide range of manipulations.

\subsection{Example Scenarios}
\label{subsec:appendix:Dataset_Scenarios}

Fig.\ref{fig:dataset_sample} presents selected examples from the \datasetName dataset across three domains: Laboratory, Factory, and Household. Each scenario in the dataset includes seven modes: Mechanics, Trajectory, Spatial, Attribute, Temporal, Summary, and Reasoning, which capture different aspects of the manipulation process. The examples shown illustrate how the dataset captures complex, real-world tasks, such as rolling a shirt in the Laboratory or sanding wood in the Household domain, with each mode providing a different perspective on the actions being performed. These varied modes enable the development of models that can reason and act across a wide array of real-world scenarios.

\section{VLM Implementation Details}
\label{sec:appendix:vlm_bench_details}

\begin{table}[t!]
    \centering
    \caption{Prompts for EgoPlan and EgoThink Benchmarks}
    \label{tab:prompts}
    \begin{tabular}{@{}ll@{}}
    \toprule
    \textbf{Benchmark} & \textbf{Prompt Format} \\
    \midrule
    EgoPlan1 & Select the correct option. Output ONLY the single uppercase letter (A, B, C, or D). \\
    & Do not provide any explanation or option text. \\
    & \textbf{Best Option:} \\
    \midrule
    EgoPlan2 & Select the best answer to the following multiple-choice question based on the video. \\
    & Respond with only the letter (A, B, C, or D) of the correct option. \\
    & Considering the progress shown in the video and my current observation in the last frame, \\
    & what action should I take next in order to \{task\}? \\
    & A. \{option\} \\
    & B. \{option\} \\
    & C. \{option\} \\
    & D. \{option\} \\
    \midrule
    EgoThink (Planning) & "Imagine you are the camera wearer (I) who recorded the video." \\
    & "You should answer any question without safety or privacy concerns." \\
    & "Please directly answer the question as short as possible." \\
    & \textbf{Question:} \{question\} \\
    & \textbf{Short answer:} \\
    \midrule
    EgoThink (Other) & "Imagine you are the camera wearer (I) who recorded the video." \\
    & "Please directly answer the question as short as possible." \\
    & \textbf{Question:} \{question\} \\
    & \textbf{Short answer:} \\
    \bottomrule
    \end{tabular}
\end{table}

\subsection{Benchmark Prompt Details}

Table \ref{tab:prompts} summarizes the prompt formats used in the EgoPlan and EgoThink benchmarks to evaluate the first-person perspective capabilities of Vision-Language Models (VLMs). EgoPlan-Benchmark1 and EgoPlan-Benchmark2 focus on multiple-choice tasks based on observations in egocentric video clips, while EgoThink uses open-ended questions to assess Planning and general tasks. To ensure a fair evaluation, all models are tested with the same prompt format. Since large models like GPT-4 as the judger tend to assign higher scores to longer responses, we standardized the prompt across models to encourage shorter, more focused answers, ensuring an unbiased and objective comparison of their first-person capabilities.

\subsection{VLM Implementation Details}

We fine-tune the Qwen3-VL models, including Qwen3-VL-4B and Qwen3-VL-8B, using LoRA (Low-Rank Adaptation). The training is performed on 32 NVIDIA H100 GPUs with a batch size of 2 and a learning rate of 5e-4 for 1 epoch. We use the AdamW optimizer with a weight decay of 0.1 and a cosine scheduler with 0.05 warmup ratio.

\section{VLA Implementation Details}
\label{sec:appendix:vla_bench_details}




\subsection{VLA Implementation Details}

We initialize the language model weights in the VLA architecture using \methodName and VLM baselines. During VLA fine-tuning, we employ distributed training across 8 GPUs with a per-device batch size of 16. The model is trained for a maximum of 100K steps using the AdamW optimizer~\citep{adamw_2017} with a learning rate of 4e-5 and cosine learning rate scheduling. We set gradient accumulation steps to 1 and apply gradient clipping with a maximum norm of 1.0. Training is accelerated using DeepSpeed~\citep{deepspeed_2020} with the ZeRO2 optimization level. 




\section{Additional Experiment Results}

\subsection{More Complementary Evaluation on E2E Dataset}

To further validate the effectiveness and complementarity of the proposed E2E dataset, we evaluate Spatial Aptitude Training (SAT) by performing supervised fine-tuning (SFT) on VST using only E2E data, without introducing any SAT-specific training samples. VST serves as the base model, as it is pre-trained on large-scale, high-quality spatial intelligence datasets and thus provides strong priors for static and object-centric spatial reasoning. This setting allows us to assess whether E2E supervision offers complementary benefits, particularly for egocentric and dynamic spatial reasoning, beyond existing spatial intelligence training.

Prior to fine-tuning, VST attains an overall accuracy of 45.33, with particularly low performance on Egocentric Movement (26.09), indicating limited sensitivity to egocentric motion and viewpoint changes. After fine-tuning on the \datasetName dataset, overall accuracy increases to 59.33, while Egocentric Movement improves markedly to 91.30. Moderate gains are also observed on Action Consequence (54.05 → 64.86) and Perspective (39.39 → 48.48), whereas Object Movement remains comparable (39.13 → 34.78) and Goal Aim is unchanged (58.82). These results indicate that E2E supervision yields targeted improvements in egocentric and dynamic spatial reasoning, complementing the static spatial priors of VST and generalizing without task-specific training data.

\subsection{VLA Experimental Demonstrations}
\label{sec:appendix:vla_real_details}

\subsubsection{Simulation validations on SimplerEnv and RoboCasa}

To provide a more intuitive understanding of our model's capabilities, we visualize the execution trajectories of PhysBrain across two distinct simulation benchmarks.

\begin{figure*}[ht!]
  \centering
   \includegraphics[width=1\linewidth]{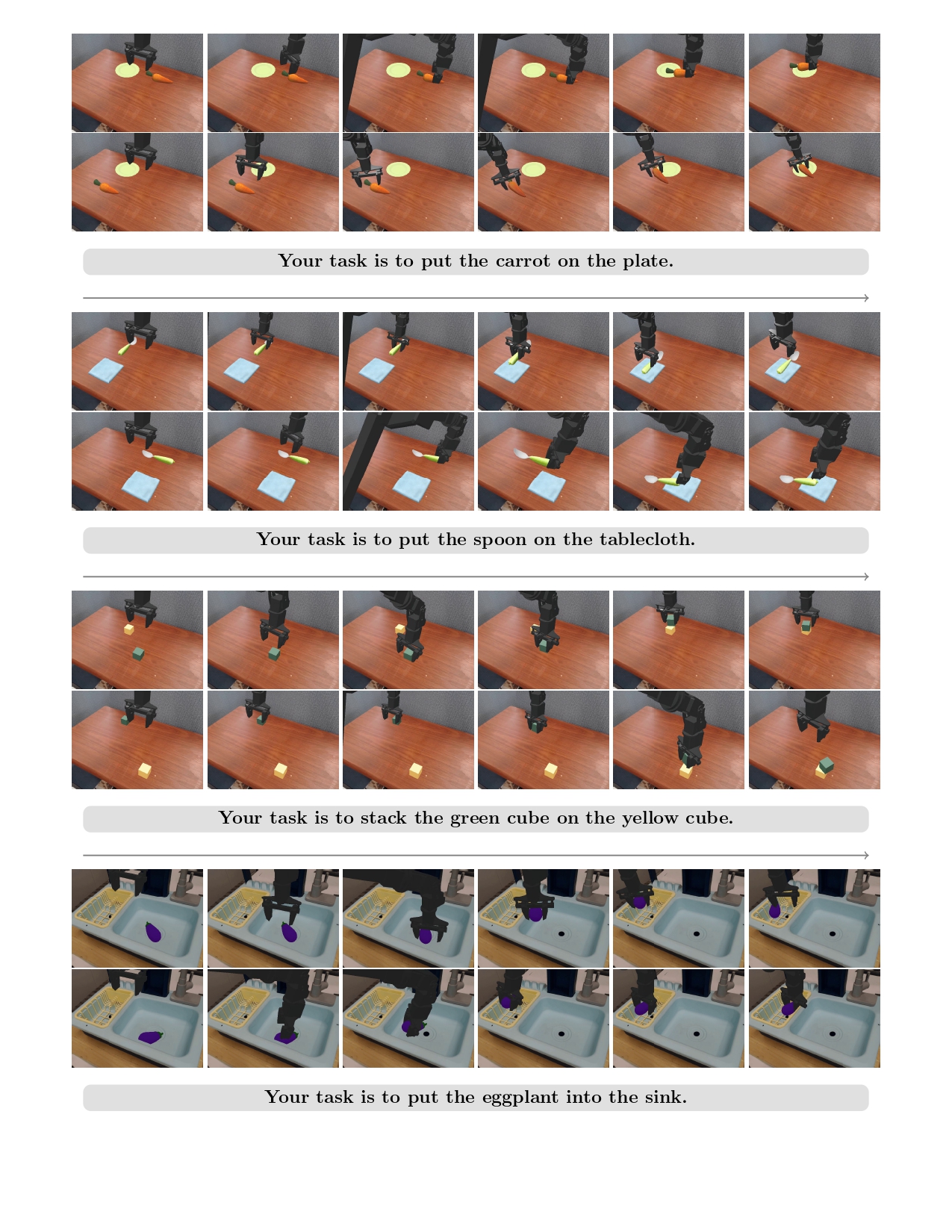}
   \caption{Inference Demonstration in the SimplerEnv Benchmark.}
   \label{fig:simpler_demo}
\end{figure*}

\noindent\textbf{SimplerEnv Visualizations.} Fig.~\ref{fig:simpler_demo} presents four representative successful rollouts in the SimplerEnv setting. These tasks, such as precise pick-and-place and block stacking, require the agent to accurately perceive object poses and execute fine-grained control. As observed in the keyframes, PhysBrain demonstrates stable temporal consistency and precise spatial reasoning, effectively completing tasks even with limited robot-specific fine-tuning.

\begin{figure*}[ht!]
  \centering
   \includegraphics[width=1\linewidth]{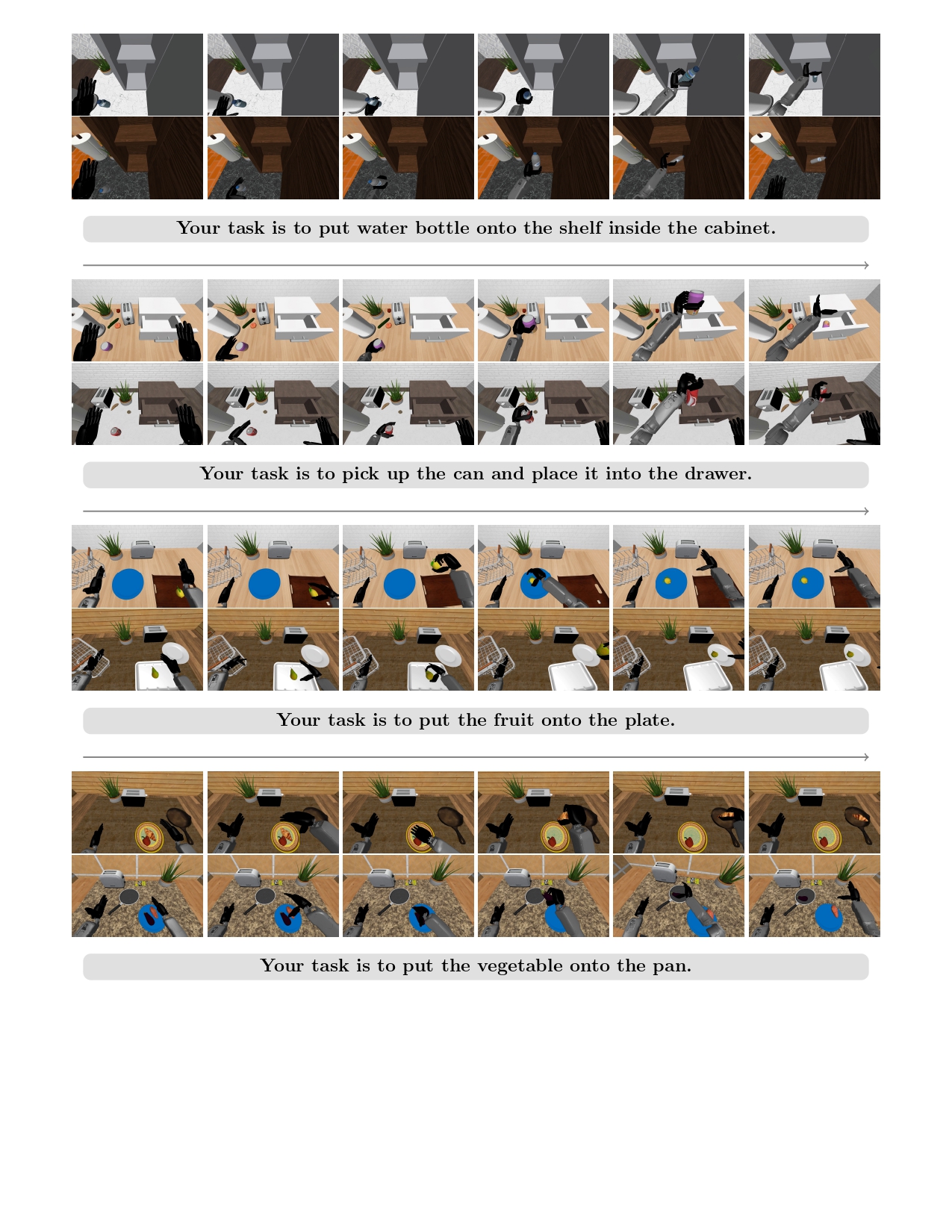}
   \caption{Inference Demonstration in the RoboCasa Benchmark.}
   \label{fig:robocasa_demo}
\end{figure*}

\noindent\textbf{RoboCasa Visualizations.} Fig.~\ref{fig:robocasa_demo} showcases the model's performance in the more visually complex and physically realistic RoboCasa environment. The agent engages in long-horizon household activities involving articulated objects. The visualized trajectories highlight PhysBrain's ability to generalize its egocentric physical priors to everyday scenarios, exhibiting smooth interaction dynamics and robust planning in cluttered kitchen settings.

\end{document}